\documentclass[11pt]{article}

\usepackage[preprint]{acl}
\usepackage{amsmath, amssymb, amsfonts}

\usepackage{times}
\usepackage{latexsym}
\usepackage{booktabs}
\usepackage{tabularx}
\usepackage{array}
\usepackage{graphicx}  
\usepackage{enumitem}  
\usepackage{dblfloatfix}
\usepackage{multirow}
\usepackage{xcolor}
\newcommand{\yl}[1]{\colorbox{yellow}{#1}}
\usepackage[utf8]{inputenc}
\usepackage{ragged2e}

\usepackage[T1]{fontenc}

\usepackage[utf8]{inputenc}

\usepackage{microtype}

\usepackage{inconsolata}

\usepackage{graphicx}

\title{Codebook-Injected Dialogue Segmentation for Multi-Utterance Constructs Annotation: LLM-Assisted and Gold-Label-Free Evaluation}

\author{
Jinsook Lee \\
Cornell University \\
\texttt{jl3369@cornell.edu}
\And
Kirk Vanacore \\
Cornell University \\
\texttt{kpv27@cornell.edu}
\And
Zhuqian Zhou \\
Cornell University \\
\texttt{zz968@cornell.edu}
\AND
Bakhtawar Ahtisham \\
Cornell University \\
\texttt{ba453@cornell.edu}
\And
Jeanine Gr{\"u}tter \\
LMU Muinich \\
\texttt{Jeanine.Gruetter@psy.lmu.de}
\And
Ren{\'e} F. Kizilcec \\
Cornell University \\
\texttt{kizilcec@cornell.edu}
}

\begin{document}
\maketitle
\begin{abstract}
Dialogue Act (DA) annotation typically treats communicative or pedagogical intent as localized to individual utterances or turns. This leads annotators to agree on the underlying action while disagreeing on segment boundaries, reducing apparent reliability. We propose \textit{codebook-injected segmentation}, which conditions boundary decisions on downstream annotation criteria, and evaluate LLM-based segmenters against standard and retrieval-augmented baselines. To assess these without gold labels, we introduce evaluation metrics for span consistency, distinctiveness, and human-AI distributional agreement. We found DA-awareness produces segments that are internally more consistent than text-only baselines. While LLMs excel at creating construct-consistent spans, coherence-based baselines remain superier at detecting global shifts in dialogue flow. Across two datasets, no single segmenter dominates: improvements in within-segment coherence frequently trade off against boundary distinctiveness and human–AI distributional agreement. These results highlight segmentation as a consequential design choice that should be optimized for downstream objectives rather than a single performance score. footnote{Code and data are available at : \href{https://github.com/National-Tutoring-Observatory/codebook-injected-segmentation}{https://github.com/National-Tutoring-Observatory/codebook-injected-segmentation}}
\end{abstract}

\section{Introduction}
High-quality labeled data is a major bottleneck for building reliable language technologies in high-stakes settings, where small annotation decisions can shift both empirical conclusions and model behavior. This has motivated extensive work on how to scale labeling while maintaining quality, including agreement measurement and pipeline design \citep{artstein2008inter}, the use of non-expert annotators to expand supervision \citep{snow2008cheap}, and crowdsourcing protocols for collecting human judgments \citep{kittur2008crowdsourcing}. However, the usefulness of a labeled dataset ultimately depends not only on reliability between annotators, but also on whether the labels faithfully capture the target constructs and can be reproduced at scale. This is particularly important for dialogue act (DA) annotation, where annotators may assign the same construct to adjacent utterances, creating boundary-driven disagreement and inflating inter-annotator disagreement \cite{ostyakova2022corpus, midgley2009dialogue}. 

DA annotation tasks encounter several design challenges, including separating semantic from pragmatic meaning, handling metadiscourse, and choosing a tagset that is both expressive and learnable \citep{verdonik2023annotating}. These challenges are not only about label choice; they also reflect a unit-of-analysis problem where coders must decide both \emph{what} the label is and \emph{where} it starts and ends. Disagreement can therefore reflect boundary misalignment, not just conceptual mismatch. Work on agreement for segmentation and ``unitizing'' continuous records similarly emphasizes that boundary placement is itself a major source of unreliability and should be modeled explicitly \citep{krippendorff2016reliability, mathet-etal-2015-unified}.

A practical response is to introduce an intermediate span level and treat segmentation as an explicit first step in the annotation workflow. Segmentation has a long history in NLP, particularly in topic and discourse analysis, and recent dialogue-specific methods improve segmentation robustness by modeling local coherence or learning representations tailored to boundary detection \citep{xing-carenini-2021-improving, xing2020improving, gao2023unsupervised}. At the same time, the expansion of supervised segmentation resources has enabled stronger learned models and more principled benchmarking of boundary definitions \citep{jiang2023superdialseg}. More recent work has explored the use of large language models (LLMs) for discourse analysis tasks, including dialogue topic segmentation, often through structured prompting that externalizes the model's boundary decisions into an interpretable output format \citep{fan-etal-2024-uncovering, das-etal-2024-s3}. In parallel, human-LLM collaborative annotation pipelines increasingly emphasize verification and oversight, acknowledging that LLM-generated annotations scale effectively only when humans can audit and intervene on intermediate decisions, including segmentation boundaries \citep{kim-etal-2024-meganno}.

This paper addresses the unit-of-analysis problem by using examples from teaching dialogue, in which the goal is to identify what instructional methods an instructor uses. These types of dialogue pose a segmentation challenge because instructors’ intent is contextual and often unfolds over multiple turns \cite{na2025llm, hennessy2016developing}. In this context, we define \textbf{tutoring moves as pedagogical dialogue acts}, functional units of instructional intent realized through a specific annotation codebook. Because these tutoring moves and instructional phases are realized as short spans rather than single utterances, boundary placement becomes an essential part of the annotation judgment. We therefore decouple boundary placement from label assignment, treating boundary decisions as an explicit, separable component of the annotation process so that multi-utterance constructs can be labeled more faithfully. The main contributions of our work are as follows:

1. We propose a codebook-injected (DA-aware) segmentation process that introduces an intermediate span level for annotation, explicitly decoupling boundary placement from label assignment so that multi-utterance constructs can be labeled more faithfully.

2. We evaluate LLM-based segmenters with both generic topic-shift prompting and DA label guidance, and benchmark them against established NLP dialogue segmentation algorithms with a retrieval-augmented variant that injects DA cues from semantically matched examples and DA labels.

3. We introduce gold-label-free evaluation criteria for segmentation in the absence of reference labels, capturing within-segment consistency, adjacent segment distinctiveness, and segment-level distributional agreement.

\section{Related Work}
\subsection{Annotating Dialogue Acts with LLMs}
Researchers increasingly utilize LLMs to scale annotation, with performance relative to human experts varying by task complexity and workflow design \citep{su2025large, yu2024assessing}. While LLMs excel in constrained settings, complex discourse annotation often requires structured pipelines to handle real-world taxonomies and multi-label cases \citep{ostyakova2023chatgpt}. Recent shifts treat LLM annotation as a \textit{process design} problem, improving robustness through multi-model ensembling and orchestration \citep{na2025llm, ahtisham2025aiannotationorchestrationevaluating}. For instance, frameworks like EduDCM improve reliability by decomposing constructs into specific components, such as acts and events, while integrating consistency checks across multiple LLMs \citep{qi2024edudcm}.

However, multiple studies have documented persistent limitations of LLMs for DA annotation. In particular, LLMs can struggle with DA classification in multi-party settings such as classroom or group discussions with multiple speakers and overlapping interactional threads \citep{qamar-etal-2025-llms}. Moreover, the complexity of goal-oriented dialogue annotation schemas (e.g., dialogue state tracking and dialogue acts) can exceed the capabilities of current models, requiring substantial human oversight \citep{labruna2023unraveling}. These analyses also emphasize concrete failure modes that matter for data quality, including hallucinated labels and instruction-following breakdowns that can degrade annotation quality relative to human coders \citep{labruna2023unraveling}. In response, some work has explored adapting models or schemas to better reflect pragmatic intent, for example, via intent-focused approaches tied to education taxonomies \citep{petukhova2025intent}, or by proposing dialogue schemas intended to be more usable for novice annotators while capturing both semantic and syntactic structure \citep{duran2022inter}.

To mitigate these issues in practice, recent work increasingly emphasizes human-in-the-loop or reliability-oriented pipelines around LLM labeling. For example, studies propose workflows that integrate verification and adjudication steps to increase reliability in semi-automated annotation \citep{10.1145/3731120.3744574, shah2025using}. At the same time, not all prompting strategies help. \citet{mohammadi-etal-2025-assessing} find that persona prompting does not improve performance on fairness-related annotation tasks (e.g., sexism). Overall, these findings suggest that using LLMs to annotate DAs at the utterance level remains challenging, and that improvements often come from workflow design, decomposition, and oversight rather than prompting alone.

\subsection{Dialogue Segmentation}
Dialogue segmentation partitions interactions into contiguous, coherent segments based on topic, task, or discourse function, with boundaries signaling meaningful shifts. Early research on monologic text utilized lexical cohesion through methods like TextTiling \cite{hearst1997texttiling} and C99 \cite{choi2000advances} to detect local drops in lexical similarity, while later Bayesian and probabilistic approaches modeled segments via latent topic distributions to infer boundaries without labeled data \cite{eisenstein2008bayesian, misra2011text}. Dialogic text introduces complexities such as lexical sparsity and pragmatic shifts, leading researchers to integrate conversational structure with lexical chains in methods like LCseg \cite{galley2003discourse} or global objectives such as minimum cuts \cite{malioutov2006minimum}. Recently, the field has transitioned toward neural and representation-based methods that capture long-range dependencies \cite{xing2020improving}, utilizing unsupervised techniques to detect coherence shifts between utterances \cite{gao2023unsupervised} or framing the task as supervised boundary prediction using section-level labeled corpora \cite{koshorek2018text}.

Segmentation is most commonly framed as a topical or discourse coherence problem, evaluated against headings, agenda items, or coarse task boundaries. Far less work treats segmentation as a deliberate pre-annotation step intended to improve the reliability and interpretability of downstream labels. Although recent work compares LLM-based segmenters to traditional methods for segmenting open-ended conversations into problem-based units aligned with predefined materials \cite{wang2024problem}, connections between segmentation choices and annotation reliability remain underexplored. This gap is especially salient in educational dialogue, where annotation targets reflect interactional strategies or pedagogical intent rather than topic alone, and boundaries may appear without strong lexical cues. 

Practically, when annotating a construct of interest that appears at the segment level, enforcing strictly local labeling can inflate apparent disagreement: annotators may share an interpretation of a multi-turn span but anchor its onset/offset differently. Utterance-aligned agreement measures can then penalize near matches while obscuring span-level convergence \citep{artstein2008inter}. We therefore frame \textbf{segmentation as a boundary setting problem} aimed at reducing ambiguity and supporting reliable segment-level labeling.

\raggedbottom

\section{Data and Context}\label{data_and_context}
To evaluate various segmentation methods, we utilize two independent educational dialogue datasets from distinct contexts, each expert-annotated with instructional move taxonomies that represent diverse pedagogical perspectives: the open-sourced TalkMoves dataset of authentic classroom discourse \cite{suresh2022talkmoves}, and a custom dataset of tutoring chat transcripts annotated using the Classroom Assessment Scoring System (CLASS) framework \cite{pianta2008class_k3}. Educational dialogue annotation provides an ideal test case for segmentation because instructional `moves' (i.e. domain-specific dialogue acts such as revoicing or scaffolding) often exist at a meso-level between a single utterance and a full session; for instance, providing an explanation may span multiple utterances and conversational turns (e.g., consider a student interjecting "ok", "yes", or "why is that" while a tutor is giving an explanation). These datasets, described in detail below and in Table \ref{tab:dataset}, capture educational interaction and are grounded in distinct conceptualizations of instructional quality, enabling evaluation across both divergent dialog settings and complementary theoretical constructs.

\begin{table}[!htbp]
\caption{Overview of datasets and label spaces.}
\raggedright
\scriptsize
\renewcommand{\arraystretch}{1.1}
\setlength{\tabcolsep}{5pt}
\begin{tabularx}{\linewidth}{@{}p{2cm}XX@{}} 
\toprule
\textbf{Feature} & \textbf{TalkMoves} & \textbf{CLASS-Annotated} \\
\midrule
\textbf{Context}
& K--12 Math Lessons \citep{suresh2022talkmoves}
& Chat-based secondary school math tutoring with annotations of instructional support dimensions \citep{pianta2008class_k3} \\
\textbf{Total Sessions}
& 63 & 30 \\
\textbf{Total Utterances}
& 31{,}263 & 1{,}881 \\
\textbf{Avg. Utterances}
& 496 & 63 \\
\textbf{Labels}
& Keeping Everyone Together; Getting students to relate to an other’s ideas; Pressing for Accuracy; Pressing for reasoning; Revoicing; Restating 
& Scaffolding; Building on Student Responses; Feedback Loop; Encouragement \\
\bottomrule
\end{tabularx}
\label{tab:dataset}
\end{table}

\subsection{TalkMoves}
The TalkMoves dataset\footnote{\url{https://github.com/SumnerLab/TalkMoves/tree/main}} is an open-source collection of mathematics classroom transcripts from authentic K-12 instructional settings, including whole-class discussions, small-group problem-solving, and online lesson contexts \citep{suresh2022talkmoves}. Transcripts were human-transcribed, segmented by speaker, and annotated at the utterance level with six pedagogically-grounded talk moves. These moves are organized into three higher-level dimensions: \emph{Learning Community}, \emph{Content Knowledge}, and \emph{Rigorous Thinking} and include: (1) \emph{Keeping Everyone Together}, \emph{Getting Students to Relate to Another's Ideas}, and \emph{Restating}; (2) \emph{Pressing for Accuracy}; and (3) \emph{Revoicing} and \emph{Pressing for Reasoning}. The original dataset includes talk-move annotations for both teacher and student utterances; in this study, we focus only on teacher talk moves. We use a subset of 63 sessions, and provide detailed move definitions in Appendix~\ref{move_definitions}.

\subsection{CLASS Annotation of Chat-based Math Tutoring Dialogue}\label{CLASS_human_annotation}
The math tutoring dialogue dataset was collected and de-identified by an online platform named Upchieve that connects volunteer tutors with students attending predominantly low-income (Title I) schools in the United States. Students initiate sessions on demand rather than following fixed schedules, and tutoring occurs through a synchronous text-based chat interface with an accompanying digital whiteboard. We focus on secondary mathematics tutoring. We randomly sampled 30 tutoring sessions for CLASS annotation by an expert who had completed formal training for applying the CLASS Instructional Support framework \citep{pianta2008class_k3, pianta2012dimensions}. While CLASS was developed to evaluate how instructors support students' conceptual understanding and reasoning in classroom settings, we worked with the trained expert to adapt and operationalize the following subdimensions applicable to a one-to-one tutoring context: (1) \emph{Feedback Loops}, (2) \emph{Scaffolding} (3) \emph{Building on Student Responses} and (4) \emph{Encouragement and Affirmation}. Detailed definitions of these four moves appear in Appendix~\ref{move_definitions}. For brevity, we refer to this dataset as CLASS-annotated throughout the paper.

\subsection{Re-Annotating Utterances with LLMs}\label{LLM_reannotation}
For both datasets, the original utterance-level DA labels were produced by expert human annotators. Following recent protocols for LLM-assisted annotation workflows \citep{ahtisham2025aiannotationorchestrationevaluating}, we additionally annotate the same transcripts with GPT-5 using LiteLLM API to obtain an LLM-generated label for each utterance. This yields two parallel rater-specific label sets per dataset, which lets us evaluate segmentation not only in terms of coherence-based criteria but also in terms of segment-level rater agreement: after aggregating utterance labels within each proposed segment into a segment-level label distribution, we measure how closely the human- and AI-derived segment distributions align under different segmentations. Full utterance-level prompts are provided in Appendix~\ref{annotation_prompt_utterance_level}.

\section{Segmentation Experiment Setup}\label{seg_exp_setup}
This section describes our task formulation and the four segmentation methods we evaluate. Two methods return boundaries using zero-shot prompting of LLMs, while the other two return boundaries using coherence-based objectives, including the Dial-Start algorithm \cite{gao2023unsupervised}. In each family, we compare a text-only variant and a DA-aware variant. For segmentation, all inputs consist only of the ordered dialogue text (utterance content) without speaker identifiers or any DA labels; DA annotations are used only for the Dial-Start DA-aware variant (as retrieval signals) and for evaluation. We chose these four methods to compare zero-shot LLM boundary proposal against a coherence-based baseline (Dial-Start), and to test whether adding DA information helps within each paradigm via matched text-only vs. DA-aware variants.

\subsection{Task Definition}
Given a dialogue as a sequence of utterances $\mathbf{u} = (u_1,\dots,u_T)$, we predict a set of boundary indices
$\mathcal{B} \subset \{1,\dots,T-1\}$, where $j\in\mathcal{B}$ denotes a boundary placed \emph{after} utterance $u_j$
(i.e., between $u_j$ and $u_{j+1}$). The boundaries induce $K=|\mathcal{B}|+1$ contiguous segments
$S_1,\dots,S_K$ such that each segment corresponds to a span in which the tutor’s pedagogical intent
(i.e., dialogue act definition) is relatively stable, and boundaries align with meaningful shifts in tutoring strategy.

\subsection{Segmentation Methods}\label{segmentation_methods}
We study dialogue segmentation as an intermediate step before downstream annotation, comparing two method families:
(i) LLM-based segmentation and (ii) unsupervised coherence-based segmentation. For each family, we evaluate a text-only variant
and a DA-aware variant that incorporates the taxonomy, enabling controlled comparisons of how explicitly conditioning on pedagogical function affects boundary placement. We hypothesize that segmentation with awareness of the annotation goal as encoded in the taxonomy (a representation of the codebook) improves the quality of subsequent annotations. 

\subsubsection{LLM Segmentation (generic prompting)}
A general-purpose LLM reads the full dialogue and outputs a set of boundary indices by marking perceived discourse or topic shifts. We ask it to segment the dialogue by outputting a JSON list of boundary indices (0-indexed turn numbers marking the last utterance in each segment).
We use GPT-5 \cite{OpenAI_GPT5_2025} and Gemini3-pro \cite{GoogleGemini3Pro} with fixed prompts and decoding settings. Prompts are provided in Appendix~\ref{sec:appendix_prompts}.

\subsubsection{LLM Segmentation (DA codebook prompting)}
To incorporate the DA information, we prompt the same LLM with the codebook (i.e., DA definitions) and instruct it to place boundaries when the pedagogical function changes, using the codebook as explicit criteria for boundary decisions. The model is prompted to refer to the definitions of DAs and add a boundary when a shift occurs. All other settings are identical to the generic prompting condition. Prompt details are in Appendix~\ref{sec:appendix_prompts} and DA
definitions are in Appendix~\ref{move_definitions}.

\subsubsection{Unsupervised Coherence Segmentation (Dial-Start)}
As a non-LLM baseline, we use Dial-Start \cite{gao2023unsupervised}, an unsupervised dialogue segmenter that proposes boundaries from local coherence drops between neighboring utterances. Dial-Start trains an utterance encoder $f(\cdot)$ using a contrastive neighboring-utterance objective that scores adjacent pairs higher than mismatched pairs. At inference time, it computes a boundary score from changes in adjacent similarity and predicts $\mathcal{B}$ using the decoding protocol in \citet{gao2023unsupervised} and specific hyperparameters are described in Appendix \ref{hyperparameter}.

\subsubsection{Unsupervised Coherence Segmentation with DA-conditioned Retrieval}
To create a DA-aware variant of the coherence baseline (Dial-Start; \citealp{gao2023unsupervised}), we add a lightweight retrieval step inspired by retrieval-augmented methods \citep{lewis2020retrieval}. Because our setting provides no gold segment boundaries, we do not train on boundary labels. Instead, we maintain a memory of expert-annotated human \emph{DA-labeles} from the same dataset (1.9k labels for TalkMoves, 301 for CLASS-annotated) to inject semantic information about DAs. For each utterance $u_i$, we retrieve the top-$K_{\text{ret}}$ semantically similar labeled utterances (by cosine similarity), aggregate their DA embeddings into a single DA vector $r_i$ (via similarity-weighted averaging), and fuse it into the utterance representation:
{\small
\begin{equation}
\hat h_i=\mathrm{norm}(h_i+\alpha r_i)    
\end{equation}
}
$\alpha$ is a fusion weight that balances the contribution of the retrieved DA embedding $r_i$ against the original utterance representation $h_i$ (see Appendix \ref{RAG} for specific values used during inference).

We then replace $h_i$ with $\hat h_i$ in Dial-Start's coherence computations. This injects codebook-aligned pedagogical signals into boundary prediction while leaving the underlying coherence objective unchanged. Details of DA-labeled memory and retrieval are provided in Appendix \ref{RAG}.

\paragraph{Compute Resources.}\label{compute_resources}
Dial-Start and its variant were run  a workstation with two NVIDIA Quadro RTX 6000 GPUs (24\,GB VRAM each). We used a single GPU and the models were implemented in PyTorch~2.9.1 (CUDA~12.8).

\subsection{Evaluation}
\begin{table}[t]
\centering
\small
\setlength{\tabcolsep}{5pt}
\begin{tabular}{@{}ll@{}}
\toprule
\textbf{Symbol} & \textbf{Meaning} \\
\midrule
$T$ & number of utterances in a dialogue \\
$K$ & number of segments \\
$S_k$ & $k$-th segment (contiguous utterances) \\
$|S_k|$ & segment length \\
$r$ & rater: Human ($H$) or AI ($A$) \\
$C$ & number of DA categories \\
$y_i^{(r)}$ & DA label for utterance $u_i$ by rater $r$ \\
$p_{k,c}^{(r)}$ & proportion of DA $c$ in segment $k$ (rater $r$) \\
$\mathbf{p}_k^{(r)}$ & DA distribution vector for segment $k$ (rater $r$) \\
$w_k$ & segment weight $|S_k|/T$ \\
$\tilde{w}_k$ & adjacent-pair weight $(|S_k|+|S_{k+1}|)/(2T)$ \\
\bottomrule
\end{tabular}
\caption{Notation used in metric definitions.}
\label{tab:metric_symbols}
\end{table}

Most segmentation evaluation methods require gold labels for segments: $P_k$ \citep{beeferman1999statistical} and WindowDiff \citep{pevzner2002critique} are two widely used metrics for evaluating text and dialogue
segmentation because they quantify boundary errors in a principled, sliding-window manner \cite{purver2011topic}. However, gold-label segments are time-intensive to produce and suffer from some of the same ambiguity issues that affect utterance-level segmentation. 

For this study, we propose a \textit{gold-label-free} evaluation approach that does not depend on reference boundaries, and instead uses downstream, distributional indicators of segment coherence and separation to compare methods. To do so, we evaluate each segmentation by converting each segment into a distribution over DA labels. We focus on the \textit{annotated DA label distributions} and measure (1) within-segment consistency, (2) between-segment distinctiveness, and (3) distribution differences between raters.

We first calculate the label distributions in each segment. A dialogue with $T$ utterances is split into $K$ contiguous segments $S_1,\dots,S_K$.
Let $r\in\{H,A\}$ denote the rater (Human vs.\ AI), and let there be $C$ DA types.
For segment $S_k$, we compute the empirical DA distribution
{\small
\begin{equation}
p_{k,c}^{(r)} = \frac{1}{|S_k|}\sum_{u_i \in S_k}\mathbb{I}[y_i^{(r)}=c],
\qquad
\sum_{c=1}^C p_{k,c}^{(r)}=1.
\end{equation}
}
We weight segments by length: $w_k = |S_k|/T$. Notation is summarized in Table~\ref{tab:metric_symbols}.

\subsubsection{Consistency within a segment}
\paragraph{Entropy (lower is better).}
Segments should have concentrated DA label distributions if they reflect a stable annotation pattern which, for our data, may represent a latent instructional method. We report a normalized entropy so that values lie in $[0,1]$ by dividing by the maximum entropy $\log_2 C$.
{\small
\begin{equation}
\begin{aligned}
H_k^{(r)} &= -\sum_{c=1}^C p_{k,c}^{(r)} \log_2 p_{k,c}^{(r)}, \\
\widetilde{H}_k^{(r)} &= \frac{H_k^{(r)}}{\log_2 C}, \\
\overline{\widetilde{H}}^{(r)} &= \sum_{k=1}^K w_k\, \widetilde{H}_k^{(r)}.
\end{aligned}
\end{equation}
}

\paragraph{Purity (higher is better).}
Purity is the share of the dominant DA label in the segment.
{\small 
\begin{equation}
\mathrm{Pur}_k^{(r)}=\max_{c} p_{k,c}^{(r)},
\qquad
\overline{\mathrm{Pur}}^{(r)} = \sum_{k=1}^K w_k\, \mathrm{Pur}_k^{(r)}.
\end{equation}
}

\subsubsection{Distinctiveness between adjacent segments}
\label{sec:distinctiveness}
Good boundaries should separate segments with meaningfully different DA profiles. We measure distinctiveness in two
complementary ways: a distributional shift between adjacent segments and a local label-transition rate at boundary positions.

\paragraph{Adjacent distribution shift via Jensen--Shannon divergence (higher is better).}
We compute the Jensen--Shannon (JS) divergence between DA label distributions of adjacent segments. For two distributions $\mathbf{p}$ and $\mathbf{q}$, JS is defined via the Kullback-Leibler divergence, $\mathrm{KL}(\mathbf{p}\|\mathbf{q}) = \sum_{i} p_i \log (p_i/q_i)$. Let $\mathbf{m}=\tfrac{1}{2}(\mathbf{p}+\mathbf{q})$; then:
{\small
\begin{equation}
\mathrm{JS}(\mathbf{p},\mathbf{q})
=\tfrac{1}{2}\mathrm{KL}(\mathbf{p}\|\mathbf{m})
+\tfrac{1}{2}\mathrm{KL}(\mathbf{q}\|\mathbf{m}).
\end{equation}
}
We aggregate adjacent distinctiveness under rater $r$ as:
{\small
\begin{equation}
\overline{\mathrm{JS}}_{\mathrm{adj}}^{(r)} = \sum_{k=1}^{K-1}\tilde{w}_k\,\mathrm{JS}\!\left(\mathbf{p}_k^{(r)},\mathbf{p}_{k+1}^{(r)}\right),
\end{equation}
}
where $\tilde{w}_k = (|S_k|+|S_{k+1}|)/2T$ weights each boundary by the proportion of the dialogue it separates.

\paragraph{Boundary change rate (higher is better).}
As a simpler boundary-level diagnostic, we compute the fraction of predicted boundaries that coincide with a change in the utterance-level DA label:
{\small
\begin{equation}
\mathrm{BCR}^{(r)}=\frac{1}{|\mathcal{B}|}\sum_{j \in \mathcal{B}}
\mathbb{I}\!\left[y_{j}^{(r)} \ne y_{j+1}^{(r)}\right].
\end{equation}
}

\subsubsection{Distributional divergence within a segment (lower is better)}
We complement IRR with a distributional agreement metric between human and AI annotations by calculating JS divergence. If a segmentation yields coherent spans, then human and AI annotations should induce similar DA distributions within the \emph{same} segments. We therefore compute
{\small
\begin{equation}
\overline{\mathrm{JS}}_{\mathrm{HA}}=
\sum_{k=1}^{K} w_k\, \mathrm{JS}\!\left(\mathbf{p}_k^{(H)},\mathbf{p}_k^{(A)}\right)
\end{equation}
}
where lower values indicate closer agreement in segment-level label distributions.

\section{Results}

Table~\ref{tab:results_main} reports average segment counts and gold-label-free quality metrics aligned with three goals: segment coherence (lower $\overline{\widetilde{H}}$ / higher $\overline{\mathrm{Pur}}$), boundary separation (higher $\overline{\mathrm{JS}}_{\mathrm{adj}}$ and $\mathrm{BCR}$), and rater alignment (lower $\overline{\mathrm{JS}}_{\mathrm{HA}}$). Qualitative examples are provided in Appendix \ref{tab:seg_example_talkmoves} and \ref{tab:seg_example_class}.


\begin{table*}[!t]
\centering
\scriptsize
\caption{Segmentation results using DA label distribution metrics, reported as mean [95\% CI]. Means are rounded to 3 decimals and CI bounds to 2 decimals. Lower $\overline{H}$ and $\overline{\mathrm{JS}}_{\mathrm{HA}}$ indicate better within-segment consistency and human-AI agreement; higher $\overline{\mathrm{JS}}_{\mathrm{adj}}$ and $\mathrm{BCR}$ indicate stronger boundary distinctiveness. Granularity is reported as the average number of segments per dialogue ($K$; mean (SD))}
\renewcommand{\arraystretch}{0.95}
\setlength{\tabcolsep}{7pt}
\begin{tabular}{@{}lcccccc@{}}
\toprule
& \multicolumn{1}{c}{\textbf{Granularity}} &
\multicolumn{2}{c}{\textbf{Consistency}} &
\multicolumn{2}{c}{\textbf{Distinctiveness}} &
\multicolumn{1}{c}{\textbf{Rater Agreement}} \\
\cmidrule(lr){2-2}\cmidrule(lr){3-4}\cmidrule(lr){5-6}\cmidrule(lr){7-7}
\textbf{Method} &
$K$ Mean (SD) &
$\overline{\widetilde{H}}$ $\downarrow$ &
$\overline{\mathrm{Pur}}$ $\uparrow$ &
$\overline{\mathrm{JS}}_{\mathrm{adj}}$ $\uparrow$ &
$\mathrm{BCR}$ $\uparrow$ &
$\overline{\mathrm{JS}}_{\mathrm{HA}}$ $\downarrow$ \\
\midrule
\multicolumn{7}{@{}l}{\textbf{CLASS-annotated}} \\
\midrule
GPT-5
& 4.90 (1.71)
& 0.349 [0.24, 0.46]
& 0.546 [0.42, 0.67]
& 0.447 [0.31, 0.59]
& 0.222 [0.14, 0.31]
& 0.424 [0.31, 0.54] \\
GPT-5 DA-aware
& 6.30 (1.97)
& \textbf{0.286} [0.18, 0.39]
& \textbf{0.570} [0.44, 0.70]
& 0.477 [0.34, 0.62]
& \textbf{0.288} [0.19, 0.38]
& 0.449 [0.33, 0.57] \\
Gemini-3-pro
& 4.47 (2.01)
& 0.384 [0.27, 0.50]
& 0.528 [0.41, 0.65]
& 0.447 [0.31, 0.59]
& 0.237 [0.13, 0.34]
& \textbf{0.407} [0.30, 0.52] \\
Gemini-3-pro DA-aware
& 4.53 (1.70)
& 0.391 [0.27, 0.51]
& 0.531 [0.41, 0.65]
& 0.435 [0.29, 0.58]
& 0.267 [0.16, 0.38]
& 0.411 [0.30, 0.52] \\
Dial-Start
& 4.60 (0.56)
& 0.303 [0.19, 0.42]
& 0.564 [0.43, 0.70]
& \textbf{0.545} [0.40, 0.69]
& 0.208 [0.12, 0.29]
& 0.459 [0.34, 0.58] \\
Dial-Start + DA-aware
& 4.50 (0.63)
& 0.319 [0.21, 0.43]
& 0.561 [0.43, 0.69]
& 0.515 [0.37, 0.66]
& 0.253 [0.15, 0.36]
& 0.484 [0.37, 0.60] \\
\midrule
\multicolumn{7}{@{}l}{\textbf{TalkMoves}} \\
\midrule
GPT-5
& 10.86 (4.87)
& 0.616 [0.55, 0.68]
& 0.659 [0.62, 0.70]
& 0.447 [0.37, 0.52]
& 0.235 [0.16, 0.31]
& \textbf{0.470} [0.44, 0.50] \\
GPT-5 DA-aware
& 12.54 (6.97)
& 0.609 [0.55, 0.67]
& 0.664 [0.63, 0.70]
& 0.470 [0.41, 0.53]
& 0.222 [0.15, 0.29]
& 0.489 [0.46, 0.52] \\
Gemini-3-pro
& 16.62 (9.09)
& 0.598 [0.53, 0.66]
& 0.662 [0.63, 0.70]
& 0.471 [0.40, 0.54]
& 0.235 [0.16, 0.31]
& 0.480 [0.45, 0.52] \\
Gemini-3-pro DA-aware
& 19.53 (10.43)
& \textbf{0.566} [0.50, 0.64]
& \textbf{0.676} [0.64, 0.71]
& \textbf{0.478} [0.41, 0.54]
& 0.252 [0.18, 0.33]
& 0.505 [0.47, 0.54] \\
Dial-Start
& 14.60 (7.27)
& 0.619 [0.56, 0.68]
& 0.640 [0.61, 0.67]
& 0.475 [0.42, 0.53]
& 0.416 [0.32, 0.52]
& 0.513 [0.48, 0.55] \\
Dial-Start + DA-aware
& 14.75 (7.18)
& 0.639 [0.58, 0.70]
& 0.633 [0.60, 0.66]
& 0.469 [0.42, 0.52]
& \textbf{0.524} [0.46, 0.59]
& 0.503 [0.47, 0.54] \\
\bottomrule
\end{tabular}
\label{tab:results_main}
\end{table*}

Two consistent patterns emerge. First, DA-awareness systematically improves consistency and distinctiveness for LLM-based segmenters but not for the coherence-based baseline. Second, LLM-based and Dial-Start segmenters optimize different and often competing objectives, with no single method dominating across metrics or datasets.

\subsection{Segmentation granularity differs modestly across methods}

Across both datasets, all methods produce a comparable number of segments per dialogue. These differences in average granularity (K) are modest relative to the high within-method variance observed in the results (Table \ref{tab:results_main}). Although DA-aware prompting leads to a moderate increase in the number of segments for LLM-based models, particularly in the TalkMoves dataset, it does not necessarily result in over-segmentation. This pattern suggests that the improvements in performance metrics are not a simple artifact of changes in granularity but instead reflect more precise boundary placement. 

\subsection{DA-awareness improves coherence for LLMs segmenters, but not for Dial-Start}
Across datasets, incorporating DA information most consistently improves within-segment coherence for LLM-based segmenters.
On CLASS-annotated, GPT-5 DA-aware yields the lowest normalized entropy and highest purity, outperforming its text-only counterpart and all non-LLM baselines. On TalkMoves the strongest coherence is achieved by Gemini-3-pro DA-aware, again improving over the text-only LLM variant (Table~\ref{tab:results_main}).

In contrast, DA-aware retrieval does not improve coherence for the coherence-based baseline. Dial-Start + DA-aware slightly degrades both entropy and purity relative to Dial-Start on both datasets, indicating that injecting DA information into a coherence objective does not make segments internally more homogeneous. Thus, DA-awareness acts as a coherence booster for LLM prompting, but not for coherence-based segmentation.

\subsection{Boundary distinctiveness depends on the method family and dataset}

Boundary quality exhibits a different pattern. On CLASS-annotated, Dial-Start achieves the strongest adjacent distinctiveness (highest $\overline{\mathrm{JS}}_{\mathrm{adj}}$), exceeding both LLM variants and Dial-Start + DA-aware. This suggests that coherence-based methods are particularly effective at separating segments with globally distinct DA distributions in this domain.

On TalkMoves, however, the strongest adjacent distinctiveness is achieved by Gemini-3-pro DA-aware, with Dial-Start close behind. Thus, when instructional moves are more varied and densely interleaved, DA-aware LLMs can rival or exceed coherence-based methods in separating distributionally distinct spans.

Local boundary contrast (BCR) further differentiates methods. On CLASS-annotated, GPT-5 DA-aware produces the highest BCR, indicating that its boundaries align most closely with local DA transitions. On TalkMoves, the highest BCR comes from Dial-Start + DA-aware, suggesting that retrieval is most effective for sharpening local, turn-level transitions in that dataset. Together, these results show that adjacent distinctiveness and local boundary contrast reward different segmentation behaviors and favor different method families.

\subsection{No single segmenter dominates: a three-way trade-off}

Across both datasets, no method simultaneously optimizes coherence, boundary separation, and human-AI distributional agreement. Improvements in coherence do not reliably translate into stronger boundary separation, and gains in either often coincide with worse human-AI alignment.

Notably, DA-aware prompting tends to increase human-AI JS divergence for LLMs, suggesting that codebook-guided boundary placement can reduce alignment with human-labeled distributions even as it improves internal coherence. In contrast, methods with stronger boundary distinctiveness often exhibit weaker rater alignment.

Overall, DA-awareness improves within-segment coherence for LLM-based segmenters, but does not consistently improve boundary quality or human–AI alignment. Across datasets, LLM-based methods tend to produce more construct-consistent spans, while coherence-based baselines more often yield sharper global boundaries.

\section{Discussion and Conclusion}

This study set out to answer two questions: whether incorporating dialogue-act (DA) information improves segmentation, and whether LLM-based segmentation is a viable alternative to coherence-based methods. DA-awareness consistently improves within-segment coherence for LLM-based segmenters, while LLM segmentation performs competitively with established coherence-based baselines. These gains are not uniform across evaluation criteria, suggesting segmentation is a multi-objective design problem rather than a single-score optimization task.

We find that improvements in within-segment coherence can conflict with boundary distinctiveness or alignment with human annotations. In pedagogical dialogue, this tension is amplified by the `unitizing' problem, where annotators often agree on instructional intent but differ on its temporal extent \citep{beeferman1999statistical, pevzner2002critique, krippendorff2016reliability, mathet-etal-2015-unified}. As a result, segmentation methods should be selected with downstream goals in mind, such as identifying extended instructional phases versus detecting fine-grained shifts.

The benefits of DA-awareness are most evident in segment coherence. Across datasets, DA-aware prompting leads LLMs to produce more homogeneous, construct-consistent spans, often outperforming text-only LLMs and, in some settings, coherence-based baselines. This suggests that LLMs can internalize codebook-level definitions of pedagogical intent and use them to guide boundary placement even without gold segmentation labels. In contrast, injecting DA information into a coherence-based objective via retrieval does not yield similar gains, indicating DA-awareness interacts differently with instruction-following models than with similarity-based criteria.

Boundary quality depends strongly on both the segmentation paradigm and the dataset. Coherence-based methods tend to place boundaries that separate globally distinct regions of dialogue, whereas LLM-based methods, especially when DA-aware, are better at carving internally consistent spans aligned with instructional constructs. Different notions of boundary success also favor different methods, and this divergence underscores that good segmentation is task-dependent.

A central empirical finding is that improvements in coherence often coincide with reduced alignment to human-labeled distributions. DA-aware prompting frequently increases human–AI divergence, revealing a tension between enforcing a formal codebook and reproducing human annotation patterns. Human annotators often rely on pragmatic judgment and contextual smoothing, implicitly tolerating boundary ambiguity in order to preserve conversational flow. By contrast, a codebook-injected LLM applies the taxonomy more literally, producing boundaries that are theoretically consistent but less sensitive to interactional nuance.

From this perspective, divergence between human and LLM segmentations should not be interpreted solely as model error. Instead, it highlights segmentation as an explicit modeling decision that shapes how constructs are operationalized. LLMs, when guided by a fixed codebook, can serve as diagnostic tools: by enforcing a stable interpretation of criteria, they can surface latent ambiguities, theoretical drift, or underspecified boundary conventions in human-annotated datasets.

Overall, our results show that segmentation should be treated as a first-class design choice and reported with the downstream use case in mind. DA-aware LLM segmentation is well suited for producing coherent spans that closely follow a formal construct definition, while coherence-based methods remain strong for detecting sharp shifts in dialogue flow. Because these approaches optimize different aspects of segmentation quality, we recommend evaluating and selecting segmenters using criteria that match the intended analysis, rather than assuming a single best method.

\section{Limitations}\label{limitations}

Our evaluation intentionally avoids gold segment boundaries, instead using distributional criteria to assess segmentation quality without reference labels. These metrics may not capture all pedagogically meaningful shifts, particularly those requiring domain expertise.
The results also depend on the underlying DA taxonomies and labeling quality, and systematic differences between human and LLM annotations can influence agreement-based measures. Esepeically for the CLASS-annotated dataset, which contains very few labeled utterances. This low density may limit the resolution of our metrics and disproportionately influence human-AI agreement measures compared to more densely labeled corpora. 
Finally, the study is limited to two datasets and a restricted set of prompting and retrieval designs, and future work is needed to assess generalization to multi-party or multimodal settings. 

\section{Ethical Considerations}\label{ethical_considerations}
We analyze de-identified tutoring dialogue datasets in accordance with our Institutional Review Board (IRB)-approved protocol (IRB0149250). TalkMoves dataset was de-identified by humans before it was open sourced and CLASS-annotated dataset was de-identified by humans by the tutoring provider and the participants consented to its use for research. We follow IRB-approved procedures for data storage and access. We use these data solely for research purposes to understand and improve dialogue annotation workflows and analyze instructional interactions. Re-identification attempts, user profiling, or any use that could enable harm to individual participants is unacceptable. Although the data are de-identified, we recognize that conversational text can still carry residual privacy risk; we therefore minimize the inclusion of verbatim excerpts, avoid reporting sensitive attributes, and present results in aggregate wherever possible. Any inferences drawn from these datasets should be interpreted cautiously, as they reflect the context and population captured by the underlying platforms and may not generalize to other settings.

\section{Acknowledgment}
We thank Doug Pietrzak and Daryl Hedley of the National Tutoring Observatory for their support with onboarding and for valuable brainstorming and technical guidance throughout this work.

\bibliography{latex/arxiv_version}

\clearpage
\appendix

\section{LLM Segmentation Prompts}
\label{sec:appendix_prompts}
Table \ref{tab:llm_prompts} summarizes the prompts used for LLM-based segmentation, comparing a zero-shot topic segmentation setting with a DA-aware setting. For the DA-aware condition, the prompt included DA (i.e. tutoring move) definitions from the codebook to guide boundary decisions.  Details of definitions are described in Appendix \ref{move_definitions}.

\begin{table*}[!htbp]
\centering
\footnotesize
\renewcommand{\arraystretch}{1.15}

\resizebox{0.9\textwidth}{!}{%
\begin{tabularx}{\textwidth}{X}
\hline

\textbf{Prompt 1: Zero-shot topic segmentation} \\
\hline
\begin{minipage}[t]{\linewidth}\vspace{2pt}
You are an expert educational discourse analyst. Your task is to segment a tutoring dialogue into coherent topics. \vspace{2pt}

\textbf{Definition:}
\begin{itemize}[leftmargin=*, itemsep=1pt, topsep=2pt]
    \item Boundary index = the 0-indexed turn number of the last utterance in a segment.
    \item Segments must be contiguous and cover the entire dialogue.
    \item Always include the final turn index as the last boundary.
\end{itemize}

\textbf{Add a boundary when:}
\begin{itemize}[leftmargin=*, itemsep=1pt, topsep=2pt]
    \item The dialogue moves to a new problem/question, or a new subtask with a different goal.
    \item The interaction shifts between social talk and lesson content.
\end{itemize}

\textbf{Do NOT add a boundary when:}
\begin{itemize}[leftmargin=*, itemsep=1pt, topsep=2pt]
    \item The tutor restates the same idea, checks understanding, or the student gives brief answers within the same topic.
\end{itemize}

\textbf{Output format} \\
Return ONLY a JSON object in the following form (no extra text, no markdown):\\
\verb|{"boundary_indices":[integer, integer, ...]}|\vspace{2pt}
\end{minipage}
\\
\hline

\textbf{Prompt 2: DA-aware segmentation} \\
\hline
\begin{minipage}[t]{\linewidth}\vspace{2pt}
You are an expert educational discourse analyst. Your task is to segment a tutoring dialogue into coherent topics. \vspace{2pt}

\textbf{Definition:}
\begin{itemize}[leftmargin=*, itemsep=1pt, topsep=2pt]
    \item Boundary index = the 0-indexed turn number of the last utterance in a segment.
    \item Segments must be contiguous and cover the entire dialogue.
    \item Always include the final turn index as the last boundary.
\end{itemize}

\textbf{Add a boundary when:}
\begin{itemize}[leftmargin=*, itemsep=1pt, topsep=2pt]
    \item The dialogue moves to a new problem/question, or a new subtask with a different goal.
    \item The interaction shifts between social talk and lesson content.
\end{itemize}

\textbf{Do NOT add a boundary when:}
\begin{itemize}[leftmargin=*, itemsep=1pt, topsep=2pt]
    \item The tutor restates the same idea, checks understanding, or the student gives brief answers within the same topic.
\end{itemize}

Use the following construct definitions to decide where meaningful boundaries occur.
Do NOT label the dialogue with these constructs; use them only to guide segmentation. \vspace{2pt}

\yl{\textbf{[ADD Move Definitions]}}

\vspace{2pt}

\textbf{Output format} \\
Return ONLY a JSON object in the following form (no extra text, no markdown):\\
\verb|{"boundary_indices":[integer, integer, ...]}| \vspace{2pt}
\end{minipage}
\\
\hline

\end{tabularx}%
}

\caption{System prompts used for LLM-based segmentation: (1) zero-shot topic segmentation and (2) DA-aware segmentation.}
\label{tab:llm_prompts}
\end{table*}

\section{Dialogue Act (DA) definitions} \label{move_definitions}
\subsection{CLASS-annotated dataset}
\paragraph{\textbf{Subcategory 1: Feedback Loops}}
Focus on the quality of feedback in the tutoring session, which is defined as the degree to which feedback advances learning and understanding and encourages student participation. A feedback loop refers to a sustained back-and-forth exchange between a teacher and a student that builds on prior turns and pushes learning forward.
\begin{itemize}
    \item Low Quality: Feedback is mostly absent, minimal, or one-sided. Exchanges are brief, superficial, and do not build on student thinking.
    \item Mid Quality: Feedback loops occur occasionally, but are inconsistent or limited in depth. Some back-and-forth exchanges involve clarifying questions or brief elaboration.
    \item High Quality: Feedback loops are frequent and often extend or deepen student understanding. The teacher regularly engages in meaningful back-and-forth exchanges that build on student thinking.
\end{itemize}

\paragraph{\textbf{Subcategory 2: Scaffolding}}
Scaffolding refers to the teacher’s use of hints, prompts, or structured support that helps students move toward successful task completion or deeper understanding. It is not however an explanation of a concept. Scaffolding enables students to perform at a higher level than they could independently by breaking down tasks, modeling thinking, or guiding problem-solving steps.
\begin{itemize}
    \item Low Quality: Students are not provided with meaningful assistance, hints, or prompting, and are often left to complete work on their own.
    \item Mid Quality: The teacher sometimes scaffolds student learning, but these interactions are often brief or shallow.
    \item High Quality: The teacher frequently scaffolds student learning, enabling students to perform at a higher level than they could independently. Support is purposeful and leads to visible progress.
\end{itemize}

\paragraph{\textbf{Subcategory 3: Building on Student Responses}}
Building on student responses refers to the teacher’s practice of expanding, instances of clarification and specific feedback, or refining what a student says to further their understanding. This includes asking targeted follow-up questions, elaborating on partial answers, or redirecting thinking in a way that pushes learning forward. It is not praise, repetition, or scaffolding.
\begin{itemize}
    \item Low Quality: The teacher gives no or very general feedback. They accept student answers and move on without clarifying or extending.
    \item Mid Quality: The teacher occasionally expands or clarifies student responses, but these exchanges are brief or shallow.
    \item High Quality: The teacher frequently builds on student responses with meaningful expansions. This includes giving specific feedback, asking clarifying or elaborative questions, and encouraging deeper reasoning.
\end{itemize}

\paragraph{\textbf{Subcategory 4: Encouragement and Affirmation}}
Encouragement and affirmation refer to the teacher’s recognition of student effort and encouragement of persistence, often through specific praise or motivational feedback that supports continued engagement in the learning process. Generic statements such as “Great,” “Okay,” “That’s correct,” or “Glad I could help” do not count as encouragement or affirmation.
\begin{itemize}
    \item Low Quality: The teacher gives little or no encouragement. Feedback is correctness-focused and does not recognize student effort or struggle.
    \item Mid Quality: The teacher occasionally encourages students with brief, general comments (e.g., “Nice try,” “Keep going”).
    \item High Quality: Encouragement targets student effort, strategies, or perseverance. It helps students persist during challenges and feel supported.
\end{itemize}

\subsection{TalkMoves dataset}
\paragraph{\textbf{Subcategory 1:  Keeping everyone together}}
Prompting students to be active listeners and orienting students to each other 

\paragraph{\textbf{Subcategory 2: Getting students to relate to another's ideas}}
Prompting students to react to what a classmate said 

\paragraph{\textbf{Subcategory 3: Restating}}
Repeating all or part of what a student said word for word

\paragraph{\textbf{Subcategory 4: Pressing for accuracy}}
Prompting students to make a mathematical contribution or use mathematical language

\paragraph{\textbf{Subcategory 5: Revoicing}}
Repeating what a student said but adding on or changing the wording
\paragraph{\textbf{Subcategory 6: Pressing for reasoning}}
Prompting students to explain, provide evidence, share their thinking behind a
decision, or connect ideas or representations

\newpage
\section{Annotation prompt at the utterance level}\label{annotation_prompt_utterance_level}

Table~\ref{tab:talkmoves_prompt_verbatim} and \ref{tab:class_feedback_prompt_verbatim} shows the prompt used to generate utterance-level labels with an LLM. We used GPT-5 to produce the annotation outputs. 

\begin{table*}[!htbp]
\centering
\scriptsize
\setlength{\tabcolsep}{6pt}
\renewcommand{\arraystretch}{1.05}
\begin{tabular}{p{0.14\textwidth} p{0.82\textwidth}}
\toprule
\textbf{Key} & \textbf{Value (verbatim)} \\
\midrule
Role &
\texttt{You are an expert educational discourse analyst. Your task is to label each teacher utterance (Speaker = T) with one Talk Move from the Open-Source Talk Moves Codebook. Use the coding definitions and examples from the manual to guide your decision. Provide a concise explanation in the 'Reasoning' field only when a Talk Move is present.}
\\

Workflow &
\texttt{Read the dialogue carefully.}\\
&\texttt{For each teacher utterance (Speaker = T), assign exactly ONE Talk Move from the Allowed Moves list.}\\
&\texttt{If an utterance could fit multiple moves, choose the one that best represents the communicative function in context.}\\
&\texttt{If no Talk Move applies (e.g., the utterance is off-topic, evaluative, unclear, or unrelated to the lesson), assign the label 'None' and leave the 'Reasoning' field empty/null. Do NOT generate reasoning for 'None' labels.}
\\

AllowedMoves &
\texttt{Keeping Everyone Together}\\
&\texttt{Getting Students to Relate to Another's Ideas}\\
&\texttt{Restating}\\
&\texttt{Pressing for Accuracy}\\
&\texttt{Revoicing}\\
&\texttt{Pressing for Reasoning}
\\

MoveDefinitions &
\texttt{"Keeping Everyone Together": "Teacher prompts students to be active listeners and orienting students to each other."}\\
&\texttt{"Getting Students to Relate to Another's Ideas": "Teacher prompts students to react to what a classmate said."}\\
&\texttt{"Restating": "Teacher repeats all or part of what a student said word for word."}\\
&\texttt{"Pressing for Accuracy": "Teacher prompts students to make a mathematical contribution or use mathematical language."}\\
&\texttt{"Revoicing": "Teacher Repeats what a student said but adding on or changing the wording."}\\
&\texttt{"Pressing for Reasoning": "Teacher prompts students to explain, provide evidence, share their thinking behind a decision, or connect ideas or representations."}
\\

MoveExamples &
\texttt{"Keeping Everyone Together": ["So x equals five dollars, right?", "It's going to be 150, right?", "Are you finished?"]}\\
&\texttt{"Getting Students to Relate to Another's Ideas": ["How do you feel about what they said?", "Does anyone understand how she solved the problem?", "Do you agree or disagree with him?"]}\\
&\texttt{"Restating": ["S: The same size and shape but it moves to a different position. \textbackslash n T: It moves to a different position. (Restating)", "S: An exponent \textbackslash n  T: Exponent. (Restating)", "S: It's four million and then the two. \textbackslash n T: Four million two. (Restating)"]}\\
&\texttt{"Pressing for Accuracy": ["What is the answer to number 2?", "How did you solve it?", "What does x stand for?"]}\\
&\texttt{"Revoicing": [" S: It had two. \textbackslash n T: So instead of one flat edge, it had two. (Revoicing)", " S: Oh, Company B. \textbackslash n T: It's Company B because that's the one that charges you \$2.00 per minute. (Revoicing)", " S: I got 2X minus Y. \textbackslash n T: It's not 2X. (Revoicing)", " S: La respuesta es siete. \textbackslash n T: The answer is seven. (Revoicing)"]}\\
&\texttt{"Pressing for Reasoning": ["Can you explain why?", "How are these ideas connected?", "Where do we see the x + 1 in the tiles?"]}
\\

RequiredOutput &
\texttt{format: JSON}\\
&\texttt{envelope: Respond with a single JSON object with a top-level field named 'records'.}\\
&\texttt{fields: ID (integer, exactly copied from the input utterance - THIS IS REQUIRED for matching); Speaker (string, exactly copied from the input utterance (usually 'T')); Turn (integer, exactly copied from the input utterance); TalkMove (one of the six Allowed Moves as a string, or null if no move applies.); Reasoning (string explanation for the chosen TalkMove, or null if no move applies.)}\\
&\texttt{example: \{"records":[\{"ID":1902,"Speaker":"T","Turn":150,"TalkMove":"Pressing for Reasoning","Reasoning":"The teacher asks students to explain whether a representation is correct, prompting justification."\},\{"ID":1905,"Speaker":"T","Turn":153,"TalkMove":null,"Reasoning":null\}]\}}
\\
\bottomrule
\end{tabular}
\caption{System prompt for utterance-level annotation (TalkMoves)}
\label{tab:talkmoves_prompt_verbatim}
\end{table*}

\begin{table*}[!t]
\centering
\scriptsize
\setlength{\tabcolsep}{6pt}
\renewcommand{\arraystretch}{1.05}
\begin{tabular}{p{0.14\textwidth} p{0.82\textwidth}}
\toprule
\textbf{Key} & \textbf{Value (verbatim)} \\
\midrule

Role &
\texttt{You are an expert in developmental and educational psychology, with advanced training using classroom observation protocols such as the Classroom Assessment Scoring System (CLASS). You will read, analyze, and annotate transcripts of teacher-student tutoring conversations for evidence of effective interactions that promote student learning.}
\\

CategoryPrompt &
\texttt{Focus on the quality of feedback in the tutoring session, which is defined as the degree to which feedback advances learning and understanding and encourages student participation.}
\\

Subcategory1\_Name &
\texttt{Feedback Loops}
\\

Subcategory1\_Prompt &
\texttt{Your task is to identify individual utterances during the tutoring session that demonstrate high-quality feedback loops. Return a value of 1 at the start of a feedback loop, leave it blank otherwise.}
\\

Subcategory1\_Definition &
\texttt{A feedback loop refers to a sustained back-and-forth exchange between a teacher and a student that builds on prior turns and pushes learning forward. Identify feedback loops that enable the student to achieve a deeper understanding of the material and concepts. In these interactions, the teacher should build upon the student's initial comment, engaging the student in sustained discussion aimed at deepening comprehension of the content.}
\\

Subcategory1\_Examples &
\texttt{1. Teacher: "What do you need to do first?" Student: "Find the diameter." Teacher: "Why start there?" Student: "It's part of the formula." Teacher: "Right---so how will you use it?"}
\\
&
\texttt{2. The teacher tells a student to include units in her answer for a math problem. Student: "Why do we have to write the units?" Teacher: "Who's going to read your answer?" Student: "You are." Teacher: "Look at the instructions. It says your answer should be clear to anyone, even someone who hasn't seen the problem." Student: "But everyone knows it's talking about centimeters." Teacher: "It might feel that way because everyone in our class is working on the same problem, but someone else reading your answer might not know what you're measuring, and the units help them understand." Student: "I guess writing the units would help someone who hasn't done the problem."}
\\
&
\texttt{3. When a student gives an incorrect answer to a problem, the teacher keeps asking follow-up questions to help the student work through their thinking until they show they understand the concept.}
\\
&
\texttt{4. A student mentions that making calculation errors is a problem in math class. The teacher asks, "What kinds of mistakes are people making in their calculations?" The student answers, "Adding instead of subtracting, or forgetting to carry numbers." The teacher asks, "Why do you think they're making these mistakes?"}
\\

Subcategory2\_Name &
\texttt{Scaffolding}
\\

Subcategory2\_Prompt &
\texttt{Your task is to identify individual utterances during the tutoring session that demonstrate high-quality Scaffolding. Return a value of 1 at the start of a Scaffolding interaction; leave it blank otherwise.}
\\

Subcategory2\_Definition &
\texttt{Scaffolding refers to the teacher's use of hints, prompts, or structured support that helps students move toward successful task completion or deeper understanding. It enables students to perform at a higher level than they could independently by breaking down tasks, modeling thinking, or guiding problem-solving steps.}
\\

Subcategory2\_Examples &
\texttt{1. The teacher notices a student who is struggling and says to her, "Let's go back and look at the relationship between the diameter and circumference of a circle."}
\\
&
\texttt{2. "If you are having difficulty with this problem, remember that it is just like the ones we did earlier."}
\\
&
\texttt{3. Teacher: "What do you need to do first? Next?" The student gives an answer. Teacher asks, "Can you explain why you think that?" or "How did you get that answer?"}
\\

Subcategory3\_Name &
\texttt{Building on student responses}
\\

Subcategory3\_Prompt &
\texttt{Your task is to identify individual utterances during the tutoring session that demonstrate high-quality building on student responses. Return a value of 1 at the start of a building on student response interaction; leave it blank otherwise.}
\\

Subcategory3\_Definition &
\texttt{Building on student responses refers to the teacher's practice of expanding, clarifying, or refining what a student says to further their understanding. This includes asking targeted follow-up questions, elaborating on partial answers, or redirecting thinking in a way that pushes learning forward.}
\\

Subcategory3\_Examples &
\texttt{1. "You're right, that's a pattern. And in this case, it repeats every three steps, which means it's a repeating pattern."}
\\
&
\texttt{2. When a student says, "The answer is 16 because I multiplied 4 and 4," the teacher responds, "Multiplying 4 by 4 does give you 16---that's one way to think about it. But you could also get 16 by adding 8 and 8. What do you notice about both of those numbers?"}
\\
&
\texttt{3. When a student says, "A system is just when everything is random," the teacher replies, "Not exactly. Remember, a system is a set of parts that work together in an organized way. Can you think of something in everyday life that works like that?"}
\\

Subcategory4\_Name &
\texttt{Encouragement and Affirmation}
\\

Subcategory4\_Prompt &
\texttt{Your task is to identify individual utterances during the tutoring session that demonstrate high-quality affirmation and encouragement. Return a value of 1 at the start of a affirmation and encouragement interaction; leave it blank otherwise.}
\\

Subcategory4\_Definition &
\texttt{Encouragement and affirmation refer to the teacher's recognition of student effort and encouragement of persistence, often through specific praise or motivational feedback that supports continued engagement in the learning process.}
\\

Subcategory4\_Examples &
\texttt{1. "That was so good! That was an incredible explanation of how you solved that problem."}
\\
&
\texttt{2. "Those are some really great ideas. You're really cranking them out!"}
\\
&
\texttt{3. "Figure it out. Don't give up. You can do it!"}
\\
&
\texttt{4. "This is a really tough topic you've picked. I am so impressed by how hard you are working. I can't wait to see the finished project!"}
\\

RequiredOutput &
\texttt{Return a value of 1 at the start of the target interaction; leave it blank otherwise.}
\\

\bottomrule
\end{tabular}
\caption{System prompt for utterance-level annotation (CLASS-annotated)}
\label{tab:class_feedback_prompt_verbatim}
\end{table*}

\newpage

\section{Hyperparameters of Dial-Start and its variant}\label{hyperparameter}
Table~\ref{tab:seg_params} summarizes the hyperparameters that control boundary selection in the coherence-based baseline. These parameters determine how boundary candidates are scored, how many boundaries can be selected per dialogue, and how close two boundaries are allowed to be.

\begin{table}[]
\centering
\small
\setlength{\tabcolsep}{5pt}
\renewcommand{\arraystretch}{1.15}
\begin{tabular}{p{0.18\linewidth} p{0.12\linewidth} p{0.62\linewidth}}
\toprule
\textbf{Parameter} & \textbf{Value} & \textbf{Role in boundary selection} \\
\midrule
\texttt{window\_size} & 2 &
Context window used when scoring each candidate boundary by comparing left/right dialogue context; larger values incorporate broader surrounding context. \\
\texttt{alpha} & 0.5 &
Threshold scale for selecting boundary candidates from depth scores:
$\mathrm{thr}=\mu+\alpha\sigma$; higher $\alpha$ yields fewer boundaries (stricter cutoff). \\
\texttt{pick\_num} & 4 &
Maximum number of boundaries returned per dialogue when \texttt{avg\_seg\_len} is unset (hard cap on selected boundary indices). \\
\texttt{avg\_seg\_len} & \texttt{None} &
If set, defines an implicit cap on boundaries via an estimated segment count
$\hat K \approx \lceil T / \texttt{avg\_seg\_len}\rceil$ (controls average segment length). \\
\texttt{min\_gap} & 3 &
Minimum separation (in utterances) enforced between consecutive boundaries to avoid trivially short adjacent segments. \\
\bottomrule
\end{tabular}
\caption{Key hyperparameters for the coherence-based segmentation baseline's boundary selection.}
\label{tab:seg_params}
\end{table}

\section{DA-labeled memory and retrieval} \label{RAG}
Let $\mathcal{D}_{\text{mem}}$ denote a pool of utterances with move labels $m\in\{1,\dots,M\}$ (e.g., human-annotated TalkMoves/CLASS-annotated labels).
We construct a memory bank of labeled embeddings
\[
\mathcal{M}=\{(h_j,m_j)\}_{j=1}^{N_{\text{mem}}}, \quad h_j=f(u_j),
\]
and, for each query utterance $u_i$ with embedding $h_i$, retrieve the top-$K_{\text{ret}}$ neighbors by cosine similarity:
\[
\{(h_{j_k},m_{j_k})\}_{k=1}^{K_{\text{ret}}}
= \textsc{TopK}_{(h,m)\in\mathcal{M}} \ \cos(h_i,h).
\]
We convert similarities into attention weights via a temperature-$\tau$ softmax,
\[
a_k = \frac{\exp(\cos(h_i,h_{j_k})/\tau)}{\sum_{\ell=1}^{K_{\text{ret}}}\exp(\cos(h_i,h_{j_\ell})/\tau)},
\]
and form an aggregated move vector using a learned move-embedding table $E\in\mathbb{R}^{M\times d}$ with row vectors $e_m$:
\[
r_i=\sum_{k=1}^{K_{\text{ret}}} a_k\, e_{m_{j_k}},\qquad
\hat h_i=\mathrm{norm}(h_i+\alpha r_i).
\]
We then replace $h_i$ with $\hat h_i$ in Dial-Start's coherence computations for boundary scoring.

\newpage

\newcolumntype{C}[1]{>{\centering\arraybackslash}p{#1}}
\newcolumntype{L}[1]{>{\RaggedRight\arraybackslash}p{#1}}
\newcolumntype{Y}{>{\RaggedRight\arraybackslash}X} 

\section{Example of Segmentation Result}
\subsection{TalkMoves}\label{example_seg_results_talkmoves}
Table \ref{tab:seg_example_talkmoves} describes an example dialogue in TalkMoves dataset. The following indices mark the turn after which each model placed a boundary:
{\small
\begin{itemize}
    \item \textbf{GPT-5}: 0, 23, 24, 60, 76
    \item \textbf{GPT-5 DA-aware}: 0, 12, 18, 23, 24, 39, 59, 76
    \item \textbf{Gemini Pro 3}: 0, 59, 76
    \item \textbf{Gemini Pro 3 DA-aware}: 0, 12, 18, 39, 58, 59, 76
    \item \textbf{Dial-Start}: 11, 44
    \item \textbf{Dial-Start + DA-aware}: 14, 44
\end{itemize}
}

\begin{table*}[!t]
\centering
\tiny
\setlength{\tabcolsep}{1pt}
\renewcommand{\arraystretch}{1.05}
\caption{An Example of Segmentation with TalkMoves dataset}
\label{tab:seg_example_talkmoves}

\begin{tabularx}{\textwidth}{@{} c Y p{0.18\textwidth} p{0.14\textwidth} c @{}}
\toprule
\textbf{Spkr} & \textbf{Utterance Content} & \textbf{Human Label} & \textbf{GPT-5 Label} & \textbf{ID}\\
\midrule
T & T and I were just chatting a little bit that its interesting for us to see all of your thinking because there are multiple ways of thinking about these problems and there are multiple ways of solving and getting the same answer &  &  & 0 \\
T & I would like to know if I look at this problem the six and a half because I know that I can do six minus four right but I cant do one twelfth minus seven twelfths it wont work because I only have one twelfth I cant take away seven twelfths &  &  & 1 \\
T & Is there a way that I could change six twelfths to help me & Keeping Everyone Together & Pressing for Accuracy & 2 \\
T & Is there a way for me to change six twelfths & Keeping Everyone Together & Pressing for Accuracy & 3 \\
T & I mean six and one twelfths sorry &  &  & 4 \\
T & How did you get seventy three twelfths & Pressing for Accuracy & Pressing for Reasoning & 5 \\
S & Well twelfth twelfths equals one whole so I did six by twelve &  &  & 6 \\
T & Then what did you do & Pressing for Accuracy & Pressing for Reasoning & 7 \\
S & Then I &  &  & 8 \\
T & Four and ten and twelve & Keeping Everyone Together &  & 9 \\
T & You know what &  &  & 10 \\
T & You did is exactly what Gavin and T were taking about over here exactly &  & Keeping Everyone Together & 11 \\
T & He changed these two into all twelfths before he subtracted them and that would work &  & Revoicing & 12 \\
T & Is there a different thing that we could do & Keeping Everyone Together & Pressing for Reasoning & 13 \\
T & Alyssa & Keeping Everyone Together &  & 14 \\
S & I did six twelfths so I did twelve twelfths and twelve twelfths plus twelve twelfths plus twelve twelfths plus twelve twelfths plus twelve twelfths and then I did plus one twelfth minus twelve twelfths minus twelve twelfths minus twelve twelfths and then I did minus seven twelfths &  &  & 15 \\
T & She broke then all into twelfths &  & Revoicing & 16 \\
S & Then I at the top I minused the four and then I took off a fifth one and then I did seven plus what equals twelve which is five so I put it at five twelfths &  &  & 17 \\
S & Then I did five twelfths plus one twelfth to get one and sixth twelfths &  &  & 18 \\
T & Okay now can I show you something &  &  & 19 \\
T & What if I treated one of these wholes what if I took a whole away from this and made it twelve twelfths & Pressing for Accuracy & Pressing for Reasoning & 20 \\
S & What you could do is you could take off two wholes and make it four and then you could minus the four &  &  & 21 \\
T & Okay so I want you to look at this &  & Keeping Everyone Together & 22 \\
T & Is five plus twelve twelfths going to give me six & Keeping Everyone Together & Pressing for Accuracy & 23 \\
T & Lindon eyes up here buddy &  & Keeping Everyone Together & 24 \\
T & Five plus twelve twelfths is going to give me six right plus I have another twelfth here so how many twelfths is that & Pressing for Accuracy & Pressing for Accuracy & 25 \\
T & Right so now watch &  & Keeping Everyone Together & 26 \\
T & I have five and thirteen twelfths now can I subtract my fractions & Keeping Everyone Together & Pressing for Accuracy & 27 \\
T & Look at that &  & Keeping Everyone Together & 28 \\
T & Okay so now I have five and thirteen twelfths minus four and seven twelfths &  &  & 29 \\
T & I can subtract those now right & Keeping Everyone Together & Pressing for Accuracy & 30 \\
T & Okay so whats my answer & Pressing for Accuracy & Pressing for Accuracy & 31 \\
T & Five minus four is & Pressing for Accuracy & Pressing for Accuracy & 32 \\
T & One and thirteen minus seven is & Pressing for Accuracy & Pressing for Accuracy & 33 \\
T & I want you to look again at what I did because I couldnt do one minus seven I had to take this and I had to trade in one of my wholes &  & Keeping Everyone Together & 34 \\
T & I had to break my whole up just like I did with that one when I had the circles here &  &  & 35 \\
T & I broke it up into fourths so I could take some away &  &  & 36 \\
T & Its the same as this &  &  & 37 \\
T & I took one of my wholes and broke it into twelve pieces and then I figured out six and one twelfth is the same as five and thirteen twelfths &  &  & 38 \\
T & I had to turn it into an improper fraction and then I subtract it &  &  & 39 \\
T & T do you have anything you want to add to that &  & Getting Students to Relate to Another's Ideas & 40 \\
T & Yes because I think about it in just a little bit of a different way &  &  & 41 \\
T & Do you want to see how I think about it &  & Keeping Everyone Together & 42 \\
T & I think you think about it like Dylan did because you shared Gavin &  &  & 43 \\
T & Just like Alyssa did &  &  & 44 \\
T & Yes because if you have twelve twelfths six times thats the same as six wholes &  &  & 45 \\
T & Yes &  &  & 46 \\
T & No &  &  & 47 \\
T & We should try it on another one &  &  & 48 \\
T & Were getting it &  &  & 49 \\
T & Wait wait wait what &  &  & 50 \\
T & Guys still need three more &  &  & 51 \\
T & I need twelve twelfths six times &  &  & 52 \\
T & And twelve times six is seventy two &  &  & 53 \\
T & Plus another one twelfth yes &  &  & 54 \\
T & So you had seventy three twelfths right &  & Revoicing & 55 \\
T & Thats how I just did it with Gavin &  &  & 56 \\
T & This is seventy three twelfths written out broken down &  &  & 57 \\
T & Thats exactly what Alyssa did too &  &  & 58 \\
T & Guys do you see the way that I would think of it versus the way T would think of it the way that Alyssa Dylan Gavin my brain thinks differently and thats okay because we would all end up at the same answer right &  & Keeping Everyone Together & 59 \\
T & I think lets go to the next one &  &  & 60 \\
T & Yes &  &  & 61 \\
T & How do you feel about doing them pair up and they talk about it with &  &  & 62 \\
T & Sure so this was the last one on here but we can do it in their journals &  &  & 63 \\
T & Do you think theyre going to need their board to do the work &  &  & 64 \\
T & There is a spot in here &  &  & 65 \\
T & Let me make sure theres enough space &  &  & 66 \\
T & Yes theres some space in here so T do you want to write page 171 up there &  &  & 67 \\
T & Up here &  &  & 68 \\
T & Yes &  &  & 69 \\
T & Its all right you can just write it &  &  & 70 \\
T & You know what I can do I can pull up the journal page on here &  &  & 71 \\
T & Heres the thing you might or might not use your board you could erase it &  &  & 72 \\
T & Were going to have to do another follow up but its a journal follow up &  &  & 73 \\
T & Oh yes page 171 &  &  & 74 \\
T & This is a journal problem and it looks like they have gave you space in your journal &  &  & 75 \\
T & If you need your whiteboard you are welcome to still use it &  &  & 76 \\
\bottomrule
\end{tabularx}
\end{table*}

\subsection{CLASS-annotated}\label{example_seg_results_CLASS}
Table \ref{tab:seg_example_class} describes an example dialogue in CLASS-annotated dataset provided by a math tutoring platform\footnote{Platform name blinded for review}. The following indices mark the turn after which each model placed a boundary:
{\small
\begin{itemize}
    \item \textbf{GPT-5}: 0, 8, 23, 30, 37
    \item \textbf{GPT-5 DA-aware}: 8, 15, 23, 27, 31, 33, 37
    \item \textbf{Gemini Pro 3}: 2, 34, 37
    \item \textbf{Gemini Pro 3 DA-aware}: 8, 30, 37
    \item \textbf{Dial-Start}: 7, 14, 22, 31
    \item \textbf{Dial-Start + DA-aware}: 2, 7, 17
\end{itemize}
}

\begin{table*}[t]
\centering
{\tiny
\setlength{\tabcolsep}{1pt}
\renewcommand{\arraystretch}{1.05}
\sloppy

\caption{An Example of Segmentation with a tutoring provider's dataset annotated with CLASS framework}
\label{tab:seg_example_class}

\begin{tabularx}{\textwidth}{@{}
C{0.06\textwidth}
Y
L{0.18\textwidth}
L{0.16\textwidth}
C{0.04\textwidth}
@{}}
\toprule
\textbf{Spkr} & \textbf{Utterance Content} & \textbf{Human Label (Move)} & \textbf{GPT-5 Label} & \textbf{ID} \\
\midrule

TEACHER & hi what can i help you with today? &  &  & 0 \\
STUDENT & hi i need help on my math homework &  &  & 1 \\
STUDENT & it's on the factor theorem and the remainder theorem. &  &  & 2 \\
STUDENT & it's as follows: &  &  & 3 \\
STUDENT & Create a polynomial $p(x)$ which has the desired characteristics. You may leave the polynomial in factored form. &  &  & 4 \\
STUDENT & (i) the zeros of $p$ are $c=2$, and $c=-2$, and $c=-1$, and $c=1$ &  &  & 5 \\
STUDENT & (ii) The leading term of $p(x)$ is $117x^2$ &  &  & 6 \\
STUDENT & There &  &  & 7 \\
STUDENT & How do i do this? &  &  & 8 \\
TEACHER & okay &  &  & 9 \\
TEACHER & is this a single problem? &  &  & 10 \\
TEACHER & since there are 4 zeroes, the leading term should be $x^4$ & Building on student responses & Scaffolding & 11 \\
STUDENT & yeah i am confused by that too &  &  & 12 \\
TEACHER & do you think it could be a typo &  & Scaffolding & 13 \\
STUDENT & i dont think so, my teacher didn't say anything about it &  &  & 14 \\
TEACHER & i think it is impossible to have 4 roots with a quadratic equation &  & Scaffolding & 15 \\
STUDENT & ill draw out what i tried &  &  & 16 \\
STUDENT & actually it was someone else's idea but neither she nor i have any idea if it's right &  &  & 17 \\
TEACHER & hm &  &  & 18 \\
TEACHER & the only problem i have with this is that it does not look like a polynomial function & Building on student responses & Scaffolding & 19 \\
STUDENT & oh okay. do you have any other ideas? im kind of lost &  &  & 20 \\
TEACHER & honestly i think it's just a typo &  &  & 21 \\
TEACHER & maybe she meant any type of function &  &  & 22 \\
TEACHER & because then this would be the answer &  & Scaffolding & 23 \\
STUDENT & what kind of function is this then? &  & Feeback Loop & 24 \\
TEACHER & a rational function &  & Feeback Loop & 25 \\
STUDENT & what is a rational function? &  & Feeback Loop & 26 \\
TEACHER & a rational function is a function with polynomials on the numerator and denominator &  & Scaffolding & 27 \\
STUDENT & so no other ideas? &  &  & 28 \\
STUDENT & if it isn't a typo i mean &  &  & 29 \\
TEACHER & i just don't think it's possible &  &  & 30 \\
STUDENT & oh okay then. thanks for trying. & Feeback Loop &  & 31 \\
TEACHER & sorry im not the teacher so i cant tell you if it should be something else &  &  & 32 \\
TEACHER & but if it doesnt fit the rules, dont be afraid to question the question & Encouragement and affirmation & Scaffolding & 33 \\
STUDENT & oh okay. i will keep that in mind. &  &  & 34 \\
STUDENT & im going to end the session now. &  &  & 35 \\
STUDENT & thanks! &  &  & 36 \\
TEACHER & okay come back if you have any other questions! & Encouragement and affirmation &  & 37 \\
\bottomrule
\end{tabularx}
}
\end{table*}

\end{document}